\documentclass[11 pt, onecolumn]{article}
\usepackage[margin=1.1in]{geometry}

\usepackage{amsmath}
\usepackage{bm}	
\usepackage{graphicx}

\usepackage{graphicx}
\usepackage{multirow}
\usepackage{mdwlist}

\usepackage{threeparttable}

\usepackage{amsfonts}
\usepackage{amsmath}
\usepackage{hyperref}
\usepackage{bbm}
\usepackage{float}
\usepackage{caption}
\usepackage{subcaption}
\usepackage{xcolor}
\usepackage{mathtools}
\usepackage{tikz}

\usetikzlibrary{shapes.geometric,arrows}
\usepackage{enumerate}
\usepackage{multicol}
\usepackage{stackrel}
\renewcommand{\)}{\right)}

\usepackage{subcaption}
\usepackage{amssymb}
\usepackage[utf8]{inputenc}
\usepackage{booktabs,lipsum}
\usepackage{hyperref}
\usepackage{algorithm}
\usepackage{algpseudocode}
\usepackage{caption}

\usepackage[numbers]{natbib}
\usepackage{longtable}

\DeclareMathAlphabet{\mathcal}{OMS}{cmsy}{m}{n}

\usepackage{graphicx}
\usepackage{epstopdf}

\begin{document}
\title{\LARGE \bf The Use of Readability Metrics in Legal Text: A Systematic Literature Review}

\author{Yu Han, 
    Aaron Ceross, 
    Jeroen H.M. Bergmann}
\newcommand*{\QEDA}{\hfill\ensuremath{\blacksquare}}%

\maketitle

\begin{abstract}
Understanding the text in legal documents can be challenging due to their complex structure and the inclusion of domain-specific jargon. Laws and regulations are often crafted in such a manner that engagement with them requires formal training, potentially leading to vastly different interpretations of the same texts. Linguistic complexity is an important contributor to the difficulties experienced by readers. Simplifying texts could enhance comprehension across a broader audience, not just among trained professionals. Various metrics have been developed to measure document readability. Therefore, we adopted a systematic review approach to examine the linguistic and readability metrics currently employed for legal and regulatory texts. A total of 3566 initial papers were screened, with 34 relevant studies found and further assessed. Our primary objective was to identify which current metrics were applied for evaluating readability within the legal field. Sixteen different metrics were identified, with the Flesch-Kincaid Grade Level being the most frequently used method. The majority of studies ($73.5\%$) were found in the domain of "informed consent forms". From the analysis, it is clear that not all legal domains are well represented in terms of readability metrics and that there is a further need to develop more consensus on which metrics should be applied for legal documents. 
\end{abstract}


\section{Introduction}
\subsection{Complexity of legislation}


Law and legislation texts can be notoriously difficult to read. The increasing number and complexity of legal documents only adds a further barrier for engagement with law by the general public \cite{rook1993laying}. In addition, overly complex regulations can lead to unnecessary economic costs to society \cite{de2022drafting}. At this point, we should probably define a bit better what is meant by complexity. According to the Oxford web dictionary, "complexity" refers to something being difficult to understand, complicated, or intricate \cite{inbook}. It also refers to something consisting of parts or elements that are not simply coordinated. The complexity of regulation may stem from various factors, including the design of regulations, the methods used to develop them, and the entities they regulate. This makes it difficult for people from all walks of life to engage with the information that is contained within these text and it can result in different interpretations between readers \cite{pagano2019reports}. The issue of complexity in legal or law texts is crucial because it can lead to confusion, loss of direction, and even malicious interpretation of the text by readers \cite{waldron1994vagueness}. Regulations or laws should provide guidance to all stakeholders and create a platform for dialogue and development while implementing the principles of law and bringing legislation into effect. Complexity issues in regulation have been researched in various fields, including general law \cite{de2022drafting}, clinical settings \cite{tegethoff2019readability,mason2018readability,paasche2003readability,arnould2021complexity}, business areas \cite{el2021voluntary}, financial regulation \cite{colliard2020measuring}, and tax \cite{smith1999readability}. According to researcher, the inability of regulation to incentivize proper behavior may have contributed to the 2008 financial crisis \cite{hakenes2014regulatory}, or at least the regulation did not manage to mitigate its consequences. Furthermore, the interpretation of legislation frequently poses challenges due to the specialized language and unique format that distinguishes it from other document types. This complexity creates an ongoing navigational and comprehension barrier for non-specialists \cite{ross1981legalities}.

\subsection{Measuring readability of legislation and regulations}

Legal law field language, which includes complex sentences archaic or apply large amount of words or expressions rarely used in other industry, or inclusion of foreign words, is often been criticized for its abstruseness \cite{williams2004legal}. People further realized the important in legal area as language as a communication tool to be understandable by most people, whether to professionals or non-professionals \cite{felsenfeld1981plain}. In the book of laws of simplicity, the researcher state that the simplest way to achieve simplicity is through thoughtful reduction and design\cite{maeda2006laws}. In legal and law field, what are been deemed as simple and useful regulation is always debatable. Readability metrics can't fix all aspects of evaluation work, for example Andrew points out that the complexity of regulation is increase not only because page number is increased, more come from major metrics calculation method has changed \cite{haldane2016complexity}. Researcher has proved readability has many limitations for example, most readability metrics are developed for children education not technical documentation\cite{redish2000readability}. 

Lawyers have “the inclination to use prepositional and other phrases in place of simple adverbs or prepositions” and they may not realize that their "stylistic choice is merely a matter of habit, because this is how lawyers traditionally speak or write" \cite{clauss2020history}. As a result, the Plain English Movement emerged in the 1980s. They were campaigning to e.g. replace old-fashioned words, avoid redundancy, reduce sentence length \cite{williams2004legal}. From this initiative, and others such as the Good Law Initiative, it became apparent that there was a need for plainer language in legal text. Simpler language would allow for better engagement with the law, whether it was for legal professionals or non-professionals. It is important to include the non-professionals, since most people in society will come in contact with it at some stage of their life \cite{felsenfeld1981plain}. The Plain English movement reinforced the importance of plain words to make legal language clearer. It highlighted the importance of using simple language to convey ideas and replacing complex expressions in legal texts. However, there is also criticism around this movement. Some argue that the plain language heritage is opposing traditional decent legal writing, signifying a new attitude and a fundamental change from previously well-established practices. Others argue that it is essentially choosing between either accuracy or clarity \cite{kimble2020flimsy}.

Legal text remain complex \cite{williams2004legal, han2024more}, despite the interest to make legal documents more accessible to the reader. People have come to realize the importance of using language that is easy to understand in order to communicate effectively with both legal professionals and non-professionals \cite{felsenfeld1981plain}. In his book "The Laws of Simplicity,"  it is suggested that the easiest way to achieve simplicity is through thoughtful reduction and design \cite{maeda2006laws}. However, what is considered simple and useful in the legal field is debatable. An ability to measure linguistic simplicity could help to reduce complexity. Readability metrics already aim to quantify how easy it is to understand a written text. However, it should be noted that the most common methods for assessing text complexity originated from techniques that were developed to determine the expected level of education required to read the text \cite{arnould2021complexity}. To what extend these metrics are applied within the legal field remains unclear.

Interpretation of (law text difference)

\subsection{Importance of readability metrics}

The development of natural language processing (NLP) techniques could enable the analysis and extraction of relevant information from legal and regulatory texts. Exploration in this field also started to include machine learning methods to assess the readability of legislative sentences \cite{curtotti2015machine}. AI could find relevant or potentially problematic clauses in millions of documents, improving efficiency and accuracy in legal analysis \cite{bommarito2021lexnlp}. Researches emphasize the need for a bridge between legal interpretation and computational challenges, highlighting the potential of NLP methods and semantic technologies in this context \cite{robaldo2019introduction}. Researcher further explores the potential of data and text mining, particularly using NLP, in the semantic analysis of legal documents \cite{waltl2018semantic}. These studies collectively underscore the transformative role of NLP in the legal domain, particularly in the analysis and extraction of relevant information from legal and regulatory texts. However, they don't show if there is common way of measuring readability nor do they reflect on which approaches have been leveraged thus far. 

While improving the readability of legislation is critical, establishing a standard method to assess and enhance readability through algorithms or readability scores remains an unresolved issue within the scientific community. In this manuscript, we acknowledge these advancements, but also recognize that there is an absence of a systematic review of the current literature.  

Consensus on how the measure readability of legal sentences before employing NLP tools can help to better compare outcomes between different studies. It can also help optimize NLP models to better handle complex legal language. A standard approach can support a much needed quality control mechanism, enabling quantitative evaluation of how changes in a text can improve readability.

The application of different readability formulas has long been a staple in evaluating the comprehensibility of texts across various domains. However, how widely these traditional formulas are applied to specialized fields remains under explored. 

Currently, no studies have systematically examined the readability metrics applied specifically to legal texts. This research aims to identify the most widely used readability metrics within the legal field and to determine the areas in which they are predominantly employed. By conducting this systematic review, we seek to offer valuable insights into the current landscape of readability assessment in legal documents and highlight areas that may require further investigation.
\section{Methodology} \label{Section: model}
\subsection*{Search Strategy}

We used the Preferred Reporting Items for Systematic Reviews and Meta-Analyses (PRISMA)~\cite{moher2009preferred} to structure the process of filtering the literature that was gathered. The literature search was conducted using three databases: Scopus, Web of Science, and IEEEXplore. We selected eight keywords and combined them with Boolean operators to obtain papers related to the legal readability or linguistic complexity. Papers published before February 2023 were included in this systematic search. Two independent reviewers read each identified relevant abstracts based on the inclusion criteria and thoroughly discussed their inclusion or exclusion. A third reviewer was used in case of disagreement. The search strategy included the following keyword combination: (complex* OR metric* OR measur*) AND (readability OR linguistic) AND (regulat* OR law OR legislation).

\subsection*{Study Selection}

The inclusion criteria consisted of: (i) it must be written in English; (ii) It must contain readability or linguistics measurements or methodology; and (iii) It must related to the legal, law or regulatory field. 

The characteristics and data extracted from each study included: (i) the year of publication; (ii) the readability measurement method used; and (iii) the domain or field of study. All eligible studies that were not accessible through library services were attempted to be obtained by contacting the corresponding author. The data extraction process was conducted independently by two reviewers to ensure accuracy and consistency. Any disagreements were resolved through discussion and consensus. A third reviewer was used again in case of any disagreement.

\subsection*{Quality Assessment}
    A modified checklist was used to evaluate the quality of each paper that was included in the final review. The quality assessment of each paper followed the questions of the adapted Specialist Unit for Review Evidence (SURE) to assist with the critical appraisal of the studies~\cite{specialist2015questions}. By leveraging the SURE approach, we aimed to minimize bias and strengthen the overall validity of our study's outcomes. Each paper included in this systematic review was checked against the items on the checklist which consisted of seven questions (see Table ~\ref{tab:criteria}). For each item, a positive answer would lead to a score of  "1", otherwise a score of "0" was assigned. For example, the question, "Does the study address a clearly focused question?" If we judged the answer to be "yes", then a score of 1 was given, otherwise a 0 was assigned. The item "Was the analysis performed by more than one researcher?" was not deemed relevant for this analysis and thus removed from the list. The quality assessment can help identify the someways that errors or biases can distort research outcomes. It provides some initial qualitative data, despite it not being a complete quality assessment. 
\begin{table}[!t]
   \centering
    \begin{tabular}{p{13.2cm}cccccccc}
    \toprule
              \textbf{Quality assessment}  \\ \midrule
            1. Does the study address a clearly focused question? \\
            2. Do the authors discuss how they decided which method to use?  \\
            3. Is there sufficient detail regarding the methods used? \\ 
            4. Are the explanations for the results plausible and coherent? \\ 
            5. Are there any potential confidentiality issues in relation to data collection?   \\ 
            6. Is any sponsorship/conflict of interest reported? \\ 
            7. Did the authors identify any limitations?  \\ \bottomrule
    \end{tabular}
   \caption{The items from the modified SURE critical appraisal checklist for systematic review.  \\ }
   \label{tab:criteria}
 \end{table}

\section{Results} 
\label{sec:results}
A total of 3,860 records were identified by searching the aforementioned databases. After removing 294 duplicates, the titles and abstracts of the remaining 3,566 papers were screened by the reviewers. Papers that were not related to the legal or regulation field were excluded from the review. Only research related to legal or law readability or complexity were kept, resulting in a total of 117 articles. Two reviewers then read the full text of each article and only kept those that contained complexity or readability metrics. After this process, 34 studies met the inclusion criteria and were further investigated. A PRISMA flowchart detailing the selection process can be found in Figure~\ref{fig:PRISMA}. The quality score of each paper is shown in Appendix 2. All included studies score full marks on the first three items.

\begin{figure}[hbt]
  \centering
  \includegraphics[width=\textwidth]{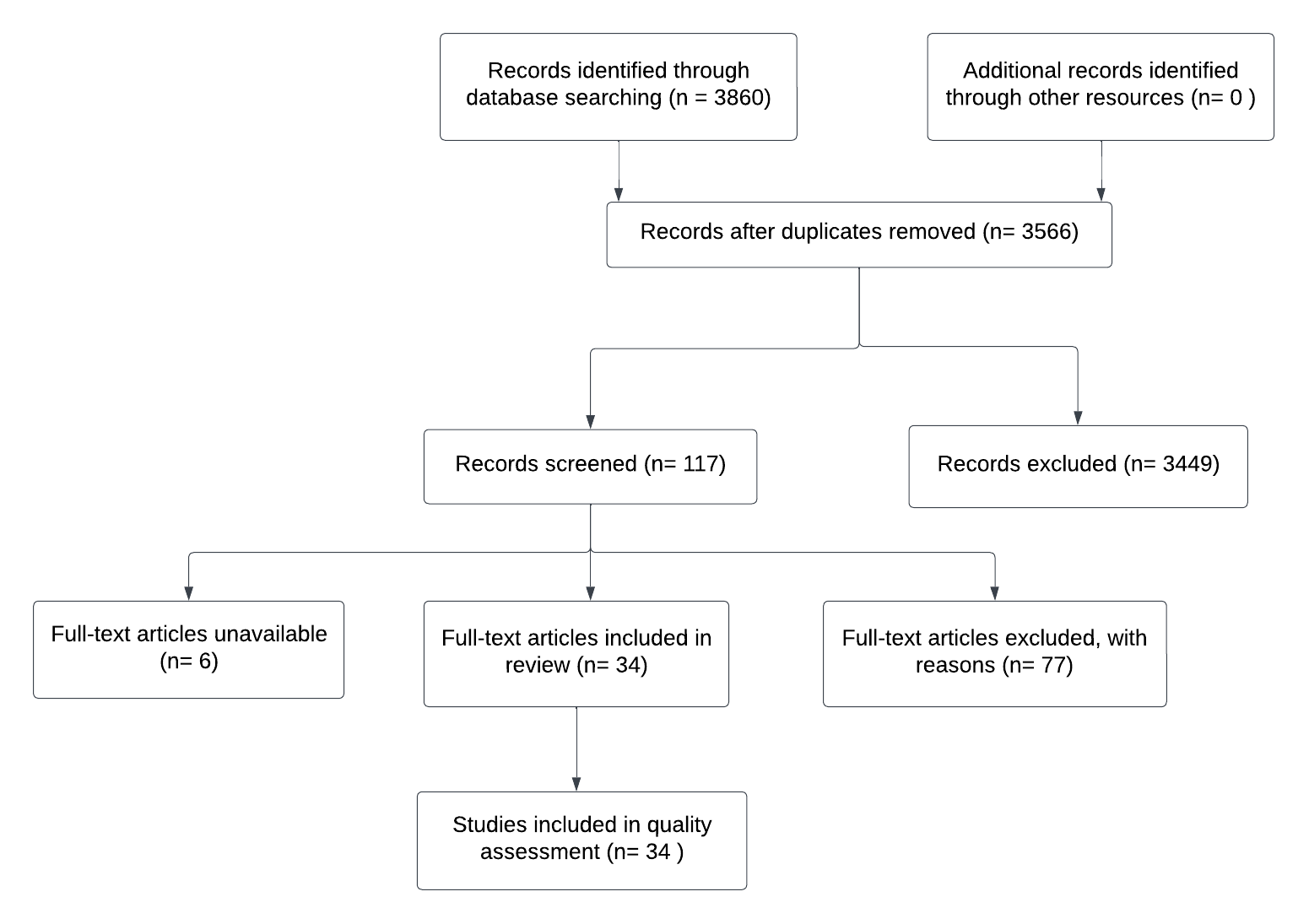}
  \caption{PRISMA flowchart of systematic review. The process consists of identification, duplicate removing, screening, and inclusion of relevant papers.}\label{fig:PRISMA}
\end{figure}

The list of the 34 papers that met the inclusion criteria are listed in Table~\ref{tab:domain}. Each paper's title, linguistic measurement used, and domain of study are included in the table.  



\begin{longtable}[!ht]{p{0.3cm}|p{4cm}|p{5cm}|p{1.5cm}}
\hline
No. &
  Title &
  Method &
  Domain \\ \hline
\endfirsthead
\endhead
1 &
  Voluntary disclosure and complexity of reporting in Egypt: the roles of profitability and earnings management\cite{el2021voluntary} &
  LIX, FOG &
  Financial \\ \hline
2 &
  Views of clinical trial participants on the readability and their understanding of informed consent documents\cite{sommers2017views} &
  Certain key word number percentage &
  Medical ICF \\ \hline 
3 &
  Variation among Consent Forms for Clinical Whole Exome Sequencing\cite{fowler2018variation} & F-KGL &
  Medical ICF  \\ \hline 
4 &
  To understand or not to understand: This is the problem\cite{tan2018understand} & Cloze Procedure, Cetinkaya and Uzun's formula, FRES, Ateşman  &
  Medical ICF  \\ \hline
5 &
  Tax law improvement in Australia and the UK: The need for a strategy for simplification\cite{james1997tax} & Consideration of rules in general terms, covering predictability, proportionality, consistency, compliance, administration, coordination and expression, etc. 
   &
  Tax \\ \hline
6 &
  Tax Law Complexity: The Impact of Style \cite{martindale1992tax} &
   FRES, F-KGL, RCE &
  Tax \\ \hline 
7 &
  A rural community's involvement in the design and usability testing of a computer-based informed consent process for the Personalized Medicine Research Project\cite{mahnke2014rural} &
  F-KGL  &
  Medical ICF \\ \hline
8 &
  Readability standards for informed-consent forms as compared with actual readability \cite{paasche2003readability}&
  F-KGL &
  Medical ICF  \\  \hline 
9 &
  The readability of informed consent forms for research studies conducted in South Africa \cite{fischer2021readability}&
  FRES, F-KGL, SMOG  &
  Medical ICF  \\  \hline
10 &
  Readability of foot and ankle consent forms in Queensland \cite{pastore2020readability} &
  F-KGL, SMOG, CLI, ARI and  Linsear Write &
  Medical ICF  \\  \hline 
11 &
  Readability of endoscopy information leaflets: Implications for informed consent \cite{mason2018readability}&
  FRES, F-KGL and SMOG &
  Medical ICF  \\  \hline 
12 &
  The Readability of Australia's Taxation Laws and Supplementary Materials: An Empirical Investigation\cite{smith1999readability} &
  FRES, F-KGL  &
  Tax law  \\  \hline
13 &
  Readability assessment of online ophthalmic patient information \cite{edmunds2013readability} &
  FRES, F-KGL, SMOG, and GFOG. &
  Medical ICF  \\  \hline
14 &
  Readability assessment of Nigerian company income tax act \cite{umar2015readability}&
  FRES, F-KGL &
  Tax Act  \\  \hline
15 &
  Readability and understandability of clinical research patient information leaflets and consent forms in Ireland and the UK: a retrospective quantitative analysis \cite{o2020readability}&
  FRES, F-KGL, SMOG, Gunning Fog, Fry, REG and Dale Chall &
  Medical ICF  \\  \hline
16 &
  Readability and content of patient information leaflets for endoscopic procedures \cite{gargoum2014readability}&
  FRES and F-KGL &
  Medical ICF \\  \hline
17 &
  Readability and Content Assessment of Informed Consent Forms for Medical Procedures in Croatia \cite{vuvcemilo2015readability}&
  SMOG  &
  Medical ICF  \\  \hline
18 &
  A randomized trial comparing concise and standard consent forms in the START trial \cite{grady2017randomized}&
  F-KGL &
  Medical ICF  \\  \hline
19 &
  Orthodontic treatment consent forms: A readability analysis\cite{meade2022orthodontic} &
  SMOG, F-KGL, FRE &
  Medical ICF  \\  \hline
20 &
  A hybrid model of complexity estimation: Evidence from Russian legal texts \cite{blinova2022hybrid}&
  Adapted F-KGL, adapted SMOG, adapted ARI, Dale–Chale, CLI &
  legal  \\  \hline
21 &
  Evaluation of length and complexity of written consent forms in English and Hebrew for participation in clinical trials authorized in one medical institution in Israel: A descriptive study \cite{ashkenazi2020evaluation}&
  FRES and F-KGL &
  Medical ICF  \\  \hline
22 &
  Evaluating the Readability of Informed Consent Forms Available Before Anaesthesia: A Comparative Study \cite{boztacs2014evaluating}&
  Gunning Fog, F-KGL and Ateşman &
  Medical ICF  \\  \hline 
23 &
  Evaluating the readability of informed consent forms used in contraceptive clinical trials\cite{rivera1992evaluating} &
  Fry Formula, Dale-Chall, SMOG formula &
  Medical ICF  \\  \hline
24 &
  Emergency medicine research consent form readability assessment\cite{mader1997emergency} &
  FRES, F-KGL and Fog  &
  Medical ICF  \\  \hline
25 &
  Consent form readability and educational levels of potential participants in mental health research\cite{christopher2007consent} &
  FRES, F-KGL, Fog , and Fry Graph &
  Medical ICF  \\  \hline
26 &
  Consent form heterogeneity in cancer trials: the cooperative group and institutional review board gap \cite{koyfman2013consent}& 
  FRES and F-KGL 
  &
  Medical ICF 
  \\  \hline
27 &
  Consent documents for oncology trials: does anybody read these things \cite{sharp2004consent}&
  FRES and Gunning Fog Index &
  Medical ICF  \\  \hline
28 &
  The Complexity of Medical Device Regulations Has Increased, as Assessed through Data-Driven Techniques \cite{arnould2021complexity}&
  Dale–Chall, ARI, CLI, Gunning Fog, F-KGL and Bog  &
  Medical Regulations \\  \hline
29 &
  Assessing the Accuracy and Readability of Online Health Information for Patients With Pancreatic Cancer \cite{storino2016assessing}&
  CLI, F-KGL, FORCAST, Fry Graph, Gunning Fog, Dale-Chall, REG, and SMOG &
  Medical ICF  \\  \hline
30 &
  Assessing readability and comprehension of informed consent materials for medical device research: A survey of informed consents from FDA's Center for Devices and Radiological Health \cite{santel2019assessing}&
  SMOG, F-KGL, FRES, and Dale-Chall &
  Medical ICF  \\  \hline
31 &
  Are informed consent forms for organ transplantation and donation too difficult to read?\cite{gordon2012informed} &
  Lexile, F-KGL, and Gunning Fog &
  Medical ICF  \\  \hline
32 &
  An analysis of the readability of patient information and consent forms used in research studies in anesthesia in Australia and New Zealand \cite{taylor2012analysis}&
  SMOG and F-KGL &
  Medical ICF \\   \hline
33 &
  Linguistic complexity of public legal information texts for young persons \cite{mamac2023linguistic} & Grammatical intricacy and lexical density & 
  Other legal\\   \hline
  34 &
  Common Contexts of Meaning in the European Legal setting opening Pandoras box \cite{ioriatti2022common}&
  Multilingual meaning problem &
  Other legal  \\   \hline 
 \caption{The title, applied method, research domain and sum of the quality score is given for each of the included studies.}
    \label{tab:domain}
\end{longtable}

An overview of the linguistic measurement methods used in the studies and their mathematical formula and other descriptions are presented in Table~\ref{tab:linguistic}.

\begin{longtable}[!ht]{p{1.5cm}|p{9cm}|p{2.7cm}}
\endfirsthead
\endhead
    \hline
         Methods & Short Description & Papers   \\ \hline
        F-KGL & The reading score formula is: 
        \[0.39 \left ( \frac{\mbox{total words}}{\mbox{total sentences}} \right ) + 11.8 \left ( \frac{\mbox{total syllables}}{\mbox{total words}} \right ) - 15.59 \]&\cite{pastore2020readability} \cite{fowler2018variation} \cite{mason2018readability} \cite{fischer2021readability}
        \cite{smith1999readability} \cite{martindale1992tax}\cite{tan2018understand} \cite{edmunds2013readability} \cite{umar2015readability} \cite{o2020readability} \cite{gargoum2014readability} \cite{grady2017randomized} \cite{meade2022orthodontic} \cite{blinova2022hybrid} \cite{ashkenazi2020evaluation} \cite{mader1997emergency} \cite{christopher2007consent} \cite{koyfman2013consent} \cite{storino2016assessing} \cite{santel2019assessing} \cite{taylor2012analysis} \cite{paasche2003readability} \cite{gordon2012informed}\\ \hline
        FRES & The reading score formula is: \[206.835 - 1.015 \left( \frac{\text{total words}}{\text{total sentences}} \right) - 84.6 \left( \frac{\text{total syllables}}{\text{total words}} \right) \]& \cite{ashkenazi2020evaluation} \cite{martindale1992tax} \cite{christopher2007consent} \cite{koyfman2013consent} \cite{sharp2004consent} \cite{arnould2021complexity} \cite{mahnke2014rural} \cite{fischer2021readability} \cite{mason2018readability} \cite{smith1999readability} \cite{edmunds2013readability} \cite{umar2015readability} \cite{o2020readability} \cite{gargoum2014readability} \cite{meade2022orthodontic} \cite{boztacs2014evaluating} \cite{storino2016assessing} \cite{santel2019assessing} \\ \hline
       SMOG &  The SMOG Readability Formula assesses text readability by counting words with three or more syllables in 30 selected sentences, then calculating a SMOG Grade. Formula is: SMOG grade = 3 + Square Root of Polysyllable Count & \cite{fischer2021readability} \cite{pastore2020readability} \cite{mason2018readability} \cite{edmunds2013readability} \cite{o2020readability} \cite{vuvcemilo2015readability} \cite{meade2022orthodontic} \cite{blinova2022hybrid} \cite{rivera1992evaluating} \cite{santel2019assessing} \cite{tegethoff2019readability}  \cite{storino2016assessing} \cite{taylor2012analysis}\\ \hline
         
      LIX & The formula is: \[ \text{LIX} = \frac{A}{B} + \frac{C \cdot 100}{A} \], where A is the number of words, B is the number of periods (defined by period, colon or capital first letter), and C is the number of long words (more than 6 letters).  &  \cite{el2021voluntary}  \\ \hline 
        
       CLI & The formula is: \[ CLI = 0.0588 \cdot L - 0.296 \cdot S - 15.8 \] 
       L is the average number of letters per 100 words and S is the average number of sentences per 100 words. & \cite{blinova2022hybrid} \cite{arnould2021complexity} \cite{pastore2020readability} \cite{storino2016assessing}\\ \hline
       
        Linsear Write & The standard Linsear Write metric $\(L_w\)$ runs on a 100-word sample:

\begin{itemize}
  \item For each "easy word," defined as words with 2 syllables or fewer, add 1 point.
  \item For each "hard word," defined as words with 3 syllables or more, add 3 points.
  \item Divide the points by the number of sentences in the 100-word sample.
  \item Adjust the provisional result $\(r\)$:
    \begin{itemize}
      \item If $\(r > 20\)$, $\(L_w = \frac{r}{2}\)$
      \item If $\(r \leq 20\)$, $\(L_w = \frac{r}{2} - 1\)$
    \end{itemize}
\end{itemize} & \cite{pastore2020readability} \\ \hline
        REG & Counting the average number of sentences and letters per 100 words &  \cite{storino2016assessing} \cite{o2020readability}\\ \hline
   Dale Chall & The reading formula is: \[ 0.1579 \left (\frac{\mbox{difficult words}}{\mbox{words}}\times 100 \right) + 0.0496 \left (\frac{\mbox{words}}{\mbox{sentences}} \right) \]
 & \cite{o2020readability} \cite{rivera1992evaluating} \cite{blinova2022hybrid} \cite{arnould2021complexity} \cite{storino2016assessing}\\ \hline
          ARI  & The formula is: \[ 4.71 \left (\frac{\mbox{characters}}{\mbox{words}} \right) + 0.5 \left (\frac{\mbox{words}}{\mbox{sentences}} \right)  - 21.43 \]
 & \cite{arnould2021complexity} \cite{pastore2020readability} \cite{blinova2022hybrid} \\ \hline
          Fog or Gunning Fog &  The formula is: \[ 0.4\left[ \left(\frac{\mbox{words}}{\mbox{sentences}}\right) + 100\left(\frac{\mbox{complex words}}{\mbox{words}}\right) \right] \]
  &  \cite{edmunds2013readability} \cite{mader1997emergency} \cite{christopher2007consent} \cite{storino2016assessing} \cite{sharp2004consent} \cite{arnould2021complexity} \\ \hline 
          Fry Graph  & Fry is a graph-based formula. It uses sentences and syllables as variables. Fry plots the text on a graph, corresponding to the score. This was originally determined by sampling excerpts from texts. & \cite{o2020readability} \cite{rivera1992evaluating} \cite{christopher2007consent} \cite{storino2016assessing}  \\ \hline
          Bog  & The Bog index consists of Sentence Bog + Word Bog – Pep Sentence Bog refers to the complexity of the sentence structure in the text. Word Bog pertains to the complexity of individual words used in the text. PEP stands for "Personal Element Percentage." It includes aspects like proper nouns (names of people, places, etc.), interest words, conversational language, and sentence variety. PEP is considered to be a factor that can make a text more engaging and easier to understand, as it introduces familiar and relatable elements.& \cite{arnould2021complexity}  \\ \hline
          FORCAST  & Grade level = \[\text{Grade level} = 20 - \left(\frac{N}{10}\right)\]
 Where N = number of single-syllable words in a 150-word sample  &  \cite{storino2016assessing} \\ \hline
          Lexile &  For an individual, a Lexile measure is typically obtained from a reading comprehension assessment or program. Measuring both reading ability and the text complexity of reading materials on the same developmental scale from 200L to 1200L. &  \cite{gordon2012informed} \\ \hline 
         Grammatical intricacy and lexical density  & Grammatical intricacy is a measure of clause complexity in texts and lexical density as the ratio of lexical items per clause. &  \cite{mamac2023linguistic} \\ \hline 
          Cloze Procedure & Similar to sentence-completion tests, the cloze method demands deletion of random words from a passage.  &  \cite{tan2018understand} \\ \hline     
    \caption{Linguistic complexity measurements methods and systematic review result.}
    \label{tab:linguistic}
\end{longtable}

 Table \ref{tab:domain2} shows the number of papers for each domain within law. The majority of papers used readability metrics in the medical field. All papers in this domain focused specifically on consent forms, apart from one which discussed "medical regulations". The next largest group of studies covered the "tax" domain, followed by "other legal" domain, whilst "Financial" had only one paper. 

\begin{table}[h]
\centering
\begin{tabular}{l r r}
\toprule
Domain & \textit{n} & \textit{pct} \\
\midrule
Medical - Informed Consent Forms & 25 & 73.5\% \\
Tax & 4 & 11.8\% \\
Legal & 2 & 5.9\% \\
Finance & 1 & 2.9\% \\
Medical - Regulations & 1 & 2.9\% \\
\bottomrule
\end{tabular}
\caption{Number and percentage of papers by legal and regulatory domain}
\label{tab:domain2}
\end{table}

 \subsection*{Three most used readability method} 
 When a document is subjected to a specific metric, it yields a calculated readability value. In general, a lower numerical value suggests better readability, implying that the document is more easily understood by a wider audience. However, there are differences between metrics and they can't be compared one to one in terms of obtained values. A more complete description of the most frequently used readability method obtained by this study is given below. 
 
 \subsection*{Flesch-Kincaid Grade Level Formula and  Flesch Reading Ease  score} 

The Flesch-Kincaid Grade Level Formula (F-KGL) is widely utilized within educational settings to estimate the grade level required to comprehend a text. This formula outputs a score corresponding to a U.S. school grade level. For instance, a F-KGL score of 10 suggests that the text is suitable for a 10th-grade student or equivalent. The F-KGL is particularly designed to indicate the complexity of English language texts and was initially developed for the U.S. Navy to improve the readability of technical manuals. 

Conversely, the Flesch Reading Ease score (FRES), often confused with the F-KGL, operates on a scale from 0 to 100, where higher scores denote easier readability. Specifically, texts scoring between 90 to 100 are considered very easy to read, appropriate for an average 11-year-old student, while scores from 0 to 30 indicate that the text is very difficult and best understood by college graduates. The FRES assesses text readability based on the number of words and sentence, as well as the syllable count per word.

To further clarify, the F-KGL outputs are based on U.S. grade levels, with higher scores indicating more difficult text. The FRES provides scores on a 0-100 scale, with higher scores reflecting easier readability.

\subsection*{SMOG} 
The Simple Measure of Gobbledygook (SMOG) Formula takes 30 sentences in a text, with 10 sentences obtained from the beginning, middle and end of the text of interest. Every word with three or more syllables in each set of sentences is then counted and the word frequency is converted into a grade level. In this case, a higher word count translates into a higher grade level. Thus, simple texts are assumed to have less words that contain three or more syllables. The approach provides a simple method to compute the readability of a certain text.  

\section*{Discussion}
In this study, we systematically searched papers and, followed PRISMA to structure the filtering to find a total of 34 papers for our final analysis. The three most commonly used readability methods identified were the Flesch-Kincaid, FRES (Flesch Reading Ease Score), and SMOG (Simple Measure of Gobbledygook) formulas. We also discovered that the majority of readability evaluations were conducted in the medical field, particularly for assessing the readability of Informed Consent Forms (ICFs). Most of the papers reviewed indicate that the readability levels of medical documents are higher than what the average person can comfortably understand. This finding highlights a potential legal issue fir medical research. Ensuring that patients can comprehend the information they receive is essential in order to obtain appropriate consent.

This study also reveals that there seems no real consensus with regards to the readability metrics that are used for legal text. Most studies appear to adopt a certain metrics for convenience, applying them without thorough examination of their suitability or effectiveness, which suggests a lack of agreement within the scientific community. This practice underscores the necessity for more rigorous discussions on readability metrics.

It is important to recognize that each readability metric is computed in (slightly) different ways. For example, the Flesch-Kincaid readability formula calculates scores based on the number of words per sentence and syllables per word, while the SMOG index specifically targets polysyllabic words, counting those with three or more syllables. This approach establishes a "threshold" value, which is present in the Flesch-Kincaid method. 

In the legal sector, it's crucial to go beyond conventional readability factors and consider unique elements such as word semantics, repetition, and sentence structure. The SMOG index, for instance, offers benefits for legal texts by analyzing polysyllabic word frequency and complex sentence structures across text samples, providing a nuanced view of text complexity typical in legal documents. However, studies show that traditional readability formulas often perform poorly in specialized fields like health information \cite{zheng2017readability}, \cite{raj2016evaluation}. The reliance on polysyllabic words in SMOG, designed initially for medical writing, may not universally apply to other languages or contexts, as it is grounded in English-language structures \cite{o2020readability}. Furthermore, medical documents often feature complex terminology and specialized knowledge, posing comprehension challenges for the general public. This suggests that less domain-specific metrics, such as the New Dale Chall, which assesses text based on the percentage of challenging or unfamiliar words, could be more suitable for broader audiences. Nonetheless, the definition of what is considered "challenging" is likely to vary across different domains. Agreement of the most suitable metrics might need to be sought for specific areas.
 
The New Dale-Chall readability formula, with its updated and modernized word list, could effectively bridge the gap in specialized fields. Yet, legal texts may still necessitate a customized word repository to accurately capture their unique complexity. In fields with specialized vocabularies, it may be essential to modify or extend this list to encompass terminology that, though potentially intricate, is commonly recognized and understood by professionals or frequent readers in those domains. This tailored approach ensures that the readability assessments are more reflective of the actual familiarity and understanding of the intended audience.


The results in Table~\ref{tab:domain} showed that there was one very specific area of the medical domain in which readability metrics were often applied. The Informed Consent Form (ICF) was often studies using such a metric. The ICF is an important tool for clinical trial teams to effectively communicate with patients. In the realm of medical practice, it's a legal obligation for professionals to secure informed consent from patients before conducting any trial, assessment, or treatment. The same principle extends to the collection of personal data for clinical trials. Consequently, signing the informed consent form signifies the patient's agreement with the propositions outlined in the patient information sheet. Ensuring informed consent mandates that participants grasp the implications of their agreement. The growing concerns about the complexity of consent documents have sparked an interest in employing quantitative methods, as highlighted by \cite{sharp2004consent}. Evidently, this field is at the forefront of incorporating readability metrics, and there is potential for other disciplines to similarly benefit from more exploration into the subject of readability.


This systematic review does not specifically determine what a good metric of readability for legal documents might be, as the review is a refection of the available research. 
Further research is needed to determine this and it probably will include considerations on accuracy,
reliability and relevance to the context of measurement. 
The metric should effectively capture the intended aspect of the analysis while being applicable across diverse scenarios \cite{dmitriev2016measuring}. 
Readability metrics might need to be expressed in terms of grade levels or other readability indices, but this might still not capture the full complexity of the text and other ways might need to be explored. Aspects such as the referencing between legal documents, the structure of the legal text, as well as the professional words applied could also be considered. 

Certain researchers have proposed other ways to measure complexity linguistically, by using for example the number of words~\cite{koster1990regulation} or pages~\cite{golub2006legal}, or legislative duration~\cite{rasmussen2013effect}, or even text’s word entropy~\cite{shannon1948mathematical}, which uses unigram tokens to measure the variety of words in a text, as this it reflects conceptual variety~\cite{katz2014predicting}. 
Since law and regulation have a special structural format, researchers have also considered to measure the structural size and element depth as an attribute of complexity~\cite{katz2014predicting}. However, this review found that the approaches that are applied currently on "legal" documents remain relatively simple. In addition, it was found that many studies only considered a single metric to draw conclusions from. Applying a wider range of metrics can provide a better reflection on the readability of a document.

Beyond the traditional readability aspects found in this study, such as average sentence, word, and syllable length, other features could be considered. Surface features including capitalization and punctuation offer basic insights into textual structure, but deeper linguistic and semantic features can provide a more nuanced understanding \cite{si2001statistical}, \cite{collins2004language}. Lexical features, for instance, encompass vocabulary richness and the type/token ratio, which reflect the diversity and complexity of language might be explored\cite{schwarm2005reading}. Additionally, parse tree depths and the frequency of subordinate clauses, which indicate the grammatical intricacy of sentences could add to the current metrics\cite{pitler2008revisiting} \cite{dell2011read}.

The Halstead complexity measures, is also something we can consider to add angles with, which originally developed for assessing software code, offer a conceptual framework that could be adapted for readability assessment. These measures consider elements such as the operational and difficulty level of the language used, which can be analogous to understanding complex legal or technical terminology. By drawing parallels to these established metrics in software engineering, we can explore potential methodologies for creating more effective readability formulas that account for domain-specific factors. These measures encompass a mechanism to categorize terms and gauge the cognitive challenge involved in comprehending a file, an aspect currently absent in prevailing readability metrics. For example, the equation for Difficulty (D) in the Halstead measures is:

\begin{equation}
D = \frac{\text{Number of Distinct Operators}}{2} \times \frac{\text{Number of Operands}}{\text{Number of Distinct Operands}}
\end{equation}

Here, operands can consist of all the words that correspond to entities (e.g., “manufacturer”), concepts (e.g., “medical device”) and values (e.g., “3 months"), whilst operators are words or combinations of words that relate to logical connections or operations, such as “and” or “all”. Jean-Edouard and Co-Pierre already drew upon this idea, successfully adapting this concept to finance regulation \cite{colliard2022measuring}. Their approach, rooted in the calculation of norms for operators and operands, presents a fresh avenue for evaluating the complexity of regulatory texts. Similarly, the Halstead method found application in medical device regulations \cite{arnould2021complexity}. This study also introduced the intriguing concept of time allocation to specific sections of text, serving as a potential marker for readability. Although this metric wouldn't yield a singular value for a given text, it generates a distribution of times associated with that text. Notably, the study revealed a relatively weak correlation between Halstead-based complexity metrics and the time individuals spent engaging with the text. This underscores the necessity to potentially explore complexity through a spectrum of metrics to enhance our comprehension of the factors influencing the readability of legal content.

We hold the perspective that readability metrics do not possess a universal solution capable of addressing all complexities associated with simplifying legal content. As mentioned, readability metrics have many limitations, as most readability metrics were developed for assessing the education of young children and might not be suitable in more technical documentation \cite{redish2000readability}. Yet, these metrics might still provide a good first step to create a more robust discussion on how we can decrease the complexity of legal documents.

When dealing with legal language, it becomes apparent that the intricacies extend beyond these conventional aspects. Elements such as word semantics, word repetition, and sentence structure play a pivotal role. In this regard, the SMOG metric holds several advantages over F-KGL. SMOG takes into consideration sentence selection from various parts of the text, thereby encompassing a more holistic perspective. Moreover, considering legal language, it is worth exploring the potential of leveraging the New Dale Chall database of words. However, it's important to note that different domains of regulation might necessitate their own repository of challenging words.

With the advancement of NLP technology, AI algorithms are increasingly harnessed to analyze and spotlight pertinent information within legal and regulatory texts in contemporary times \cite{han2024transforming}. Notably, AI machinery can sift through countless documents to pinpoint relevant or potentially problematic clauses—a boon of this era. The construction and conveyance of the original text wield an impact on fostering improved interaction with AI machines, thereby yielding enhanced outcomes in the long term \cite{bommarito2021lexnlp}. However, the existing metrics have yet to catch up with this technological progress, leading to a gap in the suitable evaluation metrics.

\section{Limitation}
This research is a systematic review that explored how often readability metrics were used for legal documents and which metrics were the most frequently applied. Our research included three distinct databases, but it needs to be acknowledged  that varying outcomes could arise based on the choice of databases. It should also be noted that this is field is rapidly evolving and searches should be reproduced to capture potential shifts within the field.

The inclusion criteria for this study was limited to the English language, potentially overlooking the incorporation of readability metrics from other languages. 

Readability assessment in other languages need to be carefully considered, as most metrics are developed upon the English language. Researchers introduce an updated measure of the readability to allow it to be used for Arabic-written texts \cite{el2021voluntary}. Other researchers took a different approach and compared English and Hebrew versions in terms of basic readability metrics \cite{ashkenazi2020evaluation}. They found that consent forms translated into English tended to be longer compared to their Hebrew counterparts. Differences between languages need to be recognised in order to create metrics that work well across regions and more research needs to be done to better understand how to create such metrics.  
\section{Conclusions} \label{Section: conclusion}

The research shows that a range of different readability metrics are currently applied within the legal domain. The most frequently used metric was the F-KCL, but there seems no clear preference for a single readability metric across the studies. When comparing various domains, the "informed consent forms" stood out with a substantial higher number of studies than any other area. In general, there seems a lack of studies in many of the legal areas. In general, this systematic review underscores a clear lack of literature concerning the assessment of complexity within legal and regulatory content for most domains. Further studies are imperative in order to better comprehend how we can enhance the clarity of legal texts and make them more accessible to a wider audience.

\section*{Abbreviation}
Flesh readability ease score = FRES \\
Flesch–Kincaid Grade Level Index = F-KGL Index \\
Simple Measure of Gobbledygook = SMOG \\
Reading Complexity Evaluation Index = RCE \\
Gunning Fog Index = GFOG \\
Automated Readability Index = ARI \\
Informed Consent Forms = ICF \\
Coleman–Liau index = CLI \\
Raygor Estimate Graph = REG \\
Informed Consent Forms = ICF 

\bibliographystyle{IEEEtran}
\bibliography{od_prediction}

\begin{thebibliography}{10}
\providecommand{\url}[1]{#1}
\csname url@samestyle\endcsname
\providecommand{\newblock}{\relax}
\providecommand{\bibinfo}[2]{#2}
\providecommand{\BIBentrySTDinterwordspacing}{\spaceskip=0pt\relax}
\providecommand{\BIBentryALTinterwordstretchfactor}{4}
\providecommand{\BIBentryALTinterwordspacing}{\spaceskip=\fontdimen2\font plus
\BIBentryALTinterwordstretchfactor\fontdimen3\font minus
  \fontdimen4\font\relax}
\providecommand{\BIBforeignlanguage}[2]{{%
\expandafter\ifx\csname l@#1\endcsname\relax
\typeout{** WARNING: IEEEtran.bst: No hyphenation pattern has been}%
\typeout{** loaded for the language `#1'. Using the pattern for}%
\typeout{** the default language instead.}%
\else
\language=\csname l@#1\endcsname
\fi
#2}}
\providecommand{\BIBdecl}{\relax}
\BIBdecl

\bibitem{rook1993laying}
L.~W. Rook, ``Laying down the law: canons for drafting complex legislation,''
  \emph{Or. L. Rev.}, vol.~72, p. 663, 1993.

\bibitem{de2022drafting}
J.~de~Lucio and J.~S. Mora-Sanguinetti, ``Drafting “better regulation”: The
  economic cost of regulatory complexity,'' \emph{Journal of Policy Modeling},
  vol.~44, no.~1, pp. 163--183, 2022.

\bibitem{inbook}
R.~Alvira, \emph{The definition of Complex and Complexity}, 09 2014, p.~12.

\bibitem{pagano2019reports}
M.~Pagano, A.~S. Serrano, and J.~Zechner, ``Reports of the advisory scientific
  committee-no 9/june 2019,'' 2019.

\bibitem{waldron1994vagueness}
J.~Waldron, ``Vagueness in law and language: Some philosophical issues,''
  \emph{Cal L. Rev.}, vol.~82, p. 509, 1994.

\bibitem{tegethoff2019readability}
D.~Tegethoff, ``Readability of information material in obstetrics,''
  \emph{Zeitschrift fur Geburtshilfe und Neonatologie}, vol. 224, no.~4, pp.
  208--216, 2019.

\bibitem{mason2018readability}
M.~C. Mason and J.~M. Williamson, ``Readability of endoscopy information
  leaflets: Implications for informed consent,'' \emph{International journal of
  clinical practice}, vol.~72, no.~5, p. e13099, 2018.

\bibitem{paasche2003readability}
M.~K. Paasche-Orlow, H.~A. Taylor, and F.~L. Brancati, ``Readability standards
  for informed-consent forms as compared with actual readability,'' \emph{New
  England journal of medicine}, vol. 348, no.~8, pp. 721--726, 2003.

\bibitem{arnould2021complexity}
A.~Arnould, R.~Hendricusdottir, and J.~Bergmann, ``The complexity of medical
  device regulations has increased, as assessed through data-driven
  techniques,'' \emph{Prosthesis}, vol.~3, no.~4, pp. 314--330, 2021.

\bibitem{el2021voluntary}
M.~M.~A. El-Din, A.~M. El-Awam, F.~M. Ibrahim, and A.~Hassanein, ``Voluntary
  disclosure and complexity of reporting in egypt: the roles of profitability
  and earnings management,'' \emph{Journal of Applied Accounting Research},
  2021.

\bibitem{colliard2020measuring}
J.-E. Colliard and C.-P. Georg, ``Measuring regulatory complexity,'' 2020.

\bibitem{smith1999readability}
D.~Smith and G.~Richardson, ``The readability of australia's taxation laws and
  supplementary materials: an empirical investigation,'' \emph{Fiscal Studies},
  vol.~20, no.~3, pp. 321--349, 1999.

\bibitem{hakenes2014regulatory}
H.~Hakenes and I.~Schnabel, ``Regulatory capture by sophistication,'' 2014.

\bibitem{ross1981legalities}
S.~M. Ross, ``On legalities and linguistics: Plain language legislation,''
  \emph{Buffalo Law Review}, vol.~30, p. 317, 1981.

\bibitem{williams2004legal}
C.~Williams \emph{et~al.}, ``Legal english and plain language: An
  introduction,'' \emph{ESP across Cultures}, vol.~1, no.~1, pp. 111--124,
  2004.

\bibitem{felsenfeld1981plain}
C.~Felsenfeld, ``Plain english movement, the plain english movement: Panel
  discussion,'' \emph{Can. Bus. LJ}, vol.~6, p. 408, 1981.

\bibitem{maeda2006laws}
J.~Maeda, \emph{The laws of simplicity}.\hskip 1em plus 0.5em minus 0.4em\relax
  MIT press, 2006.

\bibitem{haldane2016complexity}
A.~Haldane and T.~Neumann, ``Complexity in regulation,'' \emph{The Palgrave
  Handbook of European Banking}, pp. 323--347, 2016.

\bibitem{redish2000readability}
J.~Redish, ``Readability formulas have even more limitations than klare
  discusses,'' \emph{ACM Journal of Computer Documentation (JCD)}, vol.~24,
  no.~3, pp. 132--137, 2000.

\bibitem{clauss2020history}
H.~B. Clauss, ``The history of the plain language movement and legal language
  and an analysis of us nuclear treaty language,'' 2020.

\bibitem{kimble2020flimsy}
J.~Kimble, ``Flimsy claims for legalese and false criticisms of plain language:
  A 30-year collection,'' \emph{Scribes J. Leg. Writing}, vol.~19, p.~1, 2020.

\bibitem{han2024more}
Y.~Han, A.~Ceross, and J.~Bergmann, ``More than red tape: exploring complexity
  in medical device regulatory affairs,'' \emph{Frontiers in Medicine},
  vol.~11, p. 1415319, 2024.

\bibitem{curtotti2015machine}
M.~Curtotti, E.~McCreath, T.~Bruce, S.~Frug, W.~Weibel, and N.~Ceynowa,
  ``Machine learning for readability of legislative sentences,'' in
  \emph{Proceedings of the 15th International Conference on Artificial
  Intelligence and Law}, 2015, pp. 53--62.

\bibitem{bommarito2021lexnlp}
M.~J. Bommarito~II, D.~M. Katz, and E.~M. Detterman, ``Lexnlp: Natural language
  processing and information extraction for legal and regulatory texts,'' in
  \emph{Research Handbook on Big Data Law}.\hskip 1em plus 0.5em minus
  0.4em\relax Edward Elgar Publishing, 2021, pp. 216--227.

\bibitem{robaldo2019introduction}
L.~Robaldo, S.~Villata, A.~Wyner, and M.~Grabmair, ``Introduction for
  artificial intelligence and law: special issue “natural language processing
  for legal texts”,'' pp. 113--115, 2019.

\bibitem{waltl2018semantic}
B.~E. Waltl, ``Semantic analysis and computational modeling of legal
  documents,'' Ph.D. dissertation, Technische Universit{\"a}t M{\"u}nchen,
  2018.

\bibitem{moher2009preferred}
D.~Moher, A.~Liberati, J.~Tetzlaff, D.~G. Altman, and P.~Group*, ``Preferred
  reporting items for systematic reviews and meta-analyses: the prisma
  statement,'' \emph{Annals of internal medicine}, vol. 151, no.~4, pp.
  264--269, 2009.

\bibitem{specialist2015questions}
S.~U. for Review Evidence~(SURE), ``Questions to assist with the critical
  appraisal of qualitative studies,'' 2015.

\bibitem{sommers2017views}
R.~Sommers, C.~Van~Staden, and F.~Steffens, ``Views of clinical trial
  participants on the readability and their understanding of informed consent
  documents,'' \emph{AJOB Empirical Bioethics}, vol.~8, no.~4, pp. 277--284,
  2017.

\bibitem{fowler2018variation}
S.~A. Fowler, C.~J. Saunders, and M.~A. Hoffman, ``Variation among consent
  forms for clinical whole exome sequencing,'' \emph{Journal of genetic
  counseling}, vol.~27, no.~1, pp. 104--114, 2018.

\bibitem{tan2018understand}
M.~N. Tan, G.~Limnili, E.~Y{\i}ld{\i}r{\i}m, and A.~D. G{\"u}ldal, ``To
  understand or not to understand: This is the problem,'' \emph{The Turkish
  Journal of Gastroenterology}, vol.~29, no.~6, p. 642, 2018.

\bibitem{james1997tax}
S.~James and I.~Wallschutzky, ``Tax law improvement in australia and the uk:
  the need for a strategy for simplification,'' \emph{Fiscal Studies}, vol.~18,
  no.~4, pp. 445--460, 1997.

\bibitem{martindale1992tax}
B.~C. Martindale, B.~S. Koch, and S.~S. Karlinsky, ``Tax law complexity: The
  impact of style,'' \emph{The Journal of Business Communication (1973)},
  vol.~29, no.~4, pp. 383--400, 1992.

\bibitem{mahnke2014rural}
A.~N. Mahnke, J.~M. Plasek, D.~G. Hoffman, N.~S. Partridge, W.~S. Foth, C.~J.
  Waudby, L.~V. Rasmussen, V.~D. McManus, and C.~A. McCarty, ``A rural
  community's involvement in the design and usability testing of a
  computer-based informed consent process for the personalized medicine
  research project,'' \emph{American Journal of Medical Genetics Part A}, vol.
  164, no.~1, pp. 129--140, 2014.

\bibitem{fischer2021readability}
A.~Fischer, W.~Venter, S.~Collins, M.~Carman, and S.~Lalla-Edward, ``The
  readability of informed consent forms for research studies conducted in south
  africa,'' \emph{South African Medical Journal}, vol. 111, no.~2, pp.
  180--183, 2021.

\bibitem{pastore2020readability}
G.~Pastore, P.~M. Frazer, A.~Mclean, T.~P. Walsh, and S.~Platt, ``Readability
  of foot and ankle consent forms in queensland,'' \emph{ANZ Journal of
  Surgery}, vol.~90, no.~12, pp. 2549--2552, 2020.

\bibitem{edmunds2013readability}
M.~R. Edmunds, R.~J. Barry, and A.~K. Denniston, ``Readability assessment of
  online ophthalmic patient information,'' \emph{JAMA ophthalmology}, vol. 131,
  no.~12, pp. 1610--1616, 2013.

\bibitem{umar2015readability}
M.~S. Umar and N.~Saad, ``Readability assessment of nigerian company income tax
  act.'' \emph{Jurnal Pengurusan}, vol.~44, 2015.

\bibitem{o2020readability}
L.~O'Sullivan, P.~Sukumar, R.~Crowley, E.~McAuliffe, and P.~Doran,
  ``Readability and understandability of clinical research patient information
  leaflets and consent forms in ireland and the uk: a retrospective
  quantitative analysis,'' \emph{BMJ open}, vol.~10, no.~9, p. e037994, 2020.

\bibitem{gargoum2014readability}
F.~S. Gargoum and S.~T. O’Keeffe, ``Readability and content of patient
  information leaflets for endoscopic procedures,'' \emph{Irish Journal of
  Medical Science (1971-)}, vol. 183, no.~3, pp. 429--432, 2014.

\bibitem{vuvcemilo2015readability}
L.~Vu{\v{c}}emilo and A.~Borove{\v{c}}ki, ``Readability and content assessment
  of informed consent forms for medical procedures in croatia,'' \emph{PloS
  one}, vol.~10, no.~9, p. e0138017, 2015.

\bibitem{grady2017randomized}
C.~Grady, G.~Touloumi, A.~S. Walker, M.~Smolskis, S.~Sharma, A.~G. Babiker,
  N.~Pantazis, J.~Tavel, E.~Florence, A.~Sanchez \emph{et~al.}, ``A randomized
  trial comparing concise and standard consent forms in the start trial,''
  \emph{PLoS One}, vol.~12, no.~4, p. e0172607, 2017.

\bibitem{meade2022orthodontic}
M.~J. Meade and C.~W. Dreyer, ``Orthodontic treatment consent forms: A
  readability analysis,'' \emph{Journal of Orthodontics}, vol.~49, no.~1, pp.
  32--38, 2022.

\bibitem{blinova2022hybrid}
O.~Blinova and N.~Tarasov, ``A hybrid model of complexity estimation: Evidence
  from russian legal texts,'' \emph{Frontiers in Artificial Intelligence},
  vol.~5, p. 248, 2022.

\bibitem{ashkenazi2020evaluation}
I.~Ashkenazi, N.~Oster, P.~Feinberg, and O.~Olsha, ``Evaluation of length and
  complexity of written consent forms in english and hebrew for participation
  in clinical trials authorized in one medical institution in israel: a
  descriptive study,'' \emph{Accountability in Research}, vol.~27, no.~3, pp.
  138--145, 2020.

\bibitem{boztacs2014evaluating}
N.~Bozta{\c{s}}, {\c{S}}.~{\"O}zbilgin, E.~{\"O}{\c{c}}men, G.~Altunta{\c{s}},
  S.~{\"O}zkarde{\c{s}}ler, V.~Hanc{\i}, and A.~G{\"u}nerli, ``Evaluating the
  readibility of informed consent forms available before anaesthesia: a
  comparative study,'' \emph{Turkish journal of anaesthesiology and
  reanimation}, vol.~42, no.~3, p. 140, 2014.

\bibitem{rivera1992evaluating}
R.~Rivera, J.~S. Reed, and D.~Menius, ``Evaluating the readability of informed
  consent forms used in contraceptive clinical trials,'' \emph{International
  Journal of Gynecology \& Obstetrics}, vol.~38, no.~3, pp. 227--230, 1992.

\bibitem{mader1997emergency}
T.~J. Mader and S.~J. Playe, ``Emergency medicine research consent form
  readability assessment,'' \emph{Annals of emergency medicine}, vol.~29,
  no.~4, pp. 534--539, 1997.

\bibitem{christopher2007consent}
P.~P. Christopher, M.~E. Foti, K.~Roy-Bujnowski, and P.~S. Appelbaum, ``Consent
  form readability and educational levels of potential participants in mental
  health research,'' \emph{Psychiatric Services}, vol.~58, no.~2, pp. 227--232,
  2007.

\bibitem{koyfman2013consent}
S.~A. Koyfman, P.~Agre, R.~Carlisle, L.~Classen, C.~Cheatham, J.~P. Finley,
  N.~Kuhrik, M.~Kuhrik, T.~K. Mangskau, J.~O’Neill \emph{et~al.}, ``Consent
  form heterogeneity in cancer trials: the cooperative group and institutional
  review board gap,'' \emph{Journal of the National Cancer Institute}, vol.
  105, no.~13, pp. 947--953, 2013.

\bibitem{sharp2004consent}
S.~M. Sharp, ``Consent documents for oncology trials: does anybody read these
  things?'' \emph{American Journal of Clinical Oncology}, vol.~27, no.~6, pp.
  570--575, 2004.

\bibitem{storino2016assessing}
A.~Storino, M.~Castillo-Angeles, A.~A. Watkins, C.~Vargas, J.~D. Mancias,
  A.~Bullock, A.~Demirjian, A.~J. Moser, and T.~S. Kent, ``Assessing the
  accuracy and readability of online health information for patients with
  pancreatic cancer,'' \emph{JAMA surgery}, vol. 151, no.~9, pp. 831--837,
  2016.

\bibitem{santel2019assessing}
F.~Santel, I.~Bah, K.~Kim, J.-A. Lin, J.~McCracken, and A.~Teme, ``Assessing
  readability and comprehension of informed consent materials for medical
  device research: A survey of informed consents from fda's center for devices
  and radiological health,'' \emph{Contemporary Clinical Trials}, vol.~85, p.
  105831, 2019.

\bibitem{gordon2012informed}
E.~J. Gordon, A.~Bergeron, G.~McNatt, J.~Friedewald, M.~M. Abecassis, and M.~S.
  Wolf, ``Are informed consent forms for organ transplantation and donation too
  difficult to read?'' \emph{Clinical transplantation}, vol.~26, no.~2, pp.
  275--283, 2012.

\bibitem{taylor2012analysis}
H.~Taylor and D.~Bramley, ``An analysis of the readability of patient
  information and consent forms used in research studies in anaesthesia in
  australia and new zealand,'' \emph{Anaesthesia and intensive care}, vol.~40,
  no.~6, pp. 995--998, 2012.

\bibitem{mamac2023linguistic}
M.~H. Mamac, ``Linguistic complexity of public legal information texts for
  young persons,'' \emph{Text \& Talk}, 2023.

\bibitem{ioriatti2022common}
E.~Ioriatti, ``Common contexts of meaning in the european legal setting:
  Opening pandora’s box?'' \emph{International Journal for the Semiotics of
  Law-Revue internationale de S{\'e}miotique juridique}, pp. 1--17, 2022.

\bibitem{zheng2017readability}
J.~Zheng and H.~Yu, ``Readability formulas and user perceptions of electronic
  health records difficulty: a corpus study,'' \emph{Journal of medical
  Internet research}, vol.~19, no.~3, p. e59, 2017.

\bibitem{raj2016evaluation}
S.~Raj, V.~L. Sharma, A.~Singh, S.~Goel \emph{et~al.}, ``Evaluation of quality
  and readability of health information websites identified through india’s
  major search engines,'' \emph{Advances in preventive medicine}, vol. 2016,
  2016.

\bibitem{dmitriev2016measuring}
P.~Dmitriev and X.~Wu, ``Measuring metrics,'' in \emph{Proceedings of the 25th
  ACM international on conference on information and knowledge management},
  2016, pp. 429--437.

\bibitem{koster1990regulation}
A.~Koster and D.~S. Parker, ``Regulation management and logic programming,''
  \emph{Software: Practice and Experience}, vol.~20, no.~1, pp. 79--107, 1990.

\bibitem{golub2006legal}
S.~Golub, ``Legal empowerment: impact and implications for the development
  community and the world bank,'' in \emph{The World Bank Legal Review, Volume
  2: Law, Equity and Development}.\hskip 1em plus 0.5em minus 0.4em\relax Brill
  Nijhoff, 2006, pp. 167--184.

\bibitem{rasmussen2013effect}
A.~Rasmussen and D.~Toshkov, ``The effect of stakeholder involvement on
  legislative duration: Consultation of external actors and legislative
  duration in the european union,'' \emph{European Union Politics}, vol.~14,
  no.~3, pp. 366--387, 2013.

\bibitem{shannon1948mathematical}
C.~E. Shannon, ``A mathematical theory of communication,'' \emph{The Bell
  system technical journal}, vol.~27, no.~3, pp. 379--423, 1948.

\bibitem{katz2014predicting}
D.~M. Katz, M.~J. Bommarito~II, and J.~Blackman, ``Predicting the behavior of
  the supreme court of the united states: A general approach,'' \emph{arXiv
  preprint arXiv:1407.6333}, 2014.

\bibitem{si2001statistical}
L.~Si and J.~Callan, ``A statistical model for scientific readability,'' in
  \emph{Proceedings of the tenth international conference on Information and
  knowledge management}, 2001, pp. 574--576.

\bibitem{collins2004language}
K.~Collins-Thompson and J.~P. Callan, ``A language modeling approach to
  predicting reading difficulty,'' in \emph{Proceedings of the human language
  technology conference of the North American chapter of the association for
  computational linguistics: HLT-NAACL 2004}, 2004, pp. 193--200.

\bibitem{schwarm2005reading}
S.~E. Schwarm and M.~Ostendorf, ``Reading level assessment using support vector
  machines and statistical language models,'' in \emph{Proceedings of the 43rd
  annual meeting of the Association for Computational Linguistics (ACL’05)},
  2005, pp. 523--530.

\bibitem{pitler2008revisiting}
E.~Pitler and A.~Nenkova, ``Revisiting readability: A unified framework for
  predicting text quality,'' in \emph{Proceedings of the 2008 conference on
  empirical methods in natural language processing}, 2008, pp. 186--195.

\bibitem{dell2011read}
F.~Dell’Orletta, S.~Montemagni, and G.~Venturi, ``Read--it: Assessing
  readability of italian texts with a view to text simplification,'' in
  \emph{Proceedings of the second workshop on speech and language processing
  for assistive technologies}, 2011, pp. 73--83.

\bibitem{colliard2022measuring}
J.-E. Colliard and C.-P. Georg, ``Measuring regulatory complexity,'' \emph{HEC
  Paris Research Paper No. FIN-2020-1358}, 2022.

\bibitem{han2024transforming}
Y.~Han and J.~Bergmann, ``Transforming medical regulations into numbers:
  Vectorizing a decade of medical device regulatory shifts in the usa, eu, and
  china,'' \emph{arXiv preprint arXiv:2411.00567}, 2024.

\end{thebibliography}
\end{document}